\documentclass[letterpaper, 10 pt, conference]{ieeeconf}  % Comment this line out if you need a4paper

\IEEEoverridecommandlockouts                              % This command is only needed if 
\usepackage{algorithm}
\usepackage{algorithmic}
\usepackage{subfigure}
\usepackage{graphicx}
\usepackage{footnote}
\usepackage{hyperref}
\usepackage{comment}
 \RequirePackage{filecontents}

 \usepackage[acronym,toc]{glossaries}
  \newacronym{ADF}{ADF}{Assumed Density Filtering}
 \newacronym{ADE}{ADE}{Average Displacement Error}
 \newacronym{AGV}{AGV}{Autonomous Ground Vehicle}
 \newacronym{FDE}{FDE}{Final Displacement Error}
 \newacronym{BB}{BB}{Bounding Box}
 \newacronym{B-GPDM}{B-GPDM}{Balanced Gaussian Process Dynamical Models}
 \newacronym{BEV}{BEV}{birds-eye view}%Bird's Eye View}
 \newacronym{CT}{CT}{Constant Turn}
 \newacronym{CV}{CV}{Constant Velocity}
 \newacronym{CP}{CP}{Constant Position}
 \newacronym{CA}{CA}{Constant Acceleration}
 \newacronym{CNN}{CNN}{Convolutional Neural Network}
  \newacronym{ConvLSTM}{ConvLSTM}{Convolutional LSTM}
  \newacronym{CS}{CS}{Correspondence to Scene}
 \newacronym{DBN}{DBN}{Dynamic Bayesian Network}
 \newacronym{DN}{DN}{Dense Network}
 \newacronym{FFA}{FFA}{Fuzzy Finite Automata}
 \newacronym{FNR}{FNR}{False Negative Rate}
 \newacronym{EKF}{EKF}{Extended Kalman Filter}
 \newacronym{GAN}{GAN}{Generative Adversarial Network}
 \newacronym{GT}{GT}{Ground Truth}
 \newacronym{GP}{GP}{Gaussian Process}
 \newacronym{GRU}{GRU}{Gated Recurrent Unit}
 \newacronym{GPDM}{GPDM}{Gaussian Process Dynamical Models}
 \newacronym{HMI}{HMI}{Human-Machine Interaction}
 \newacronym{IMM}{IMM}{Interacting Multiple Model}
 \newacronym{ITS}{ITS}{Intelligent Transportation System}
 \newacronym{IMM-EKF}{IMM-EKF}{Interacting Multiple Model Extended Kalman Filter}
  \newacronym{ISPRS}{ISPRS}{International Society for Photogrammetry and Remote Sensing}
 \newacronym{KF}{KF}{Kalman Filter}
 \newacronym{LDS}{LDS}{Linear Dynamical System}
 \newacronym{LRM}{LRM}{Mobile Robotics Laboratory}
 \newacronym{LDCRF}{LDCRF}{Latent Dynamic Conditional Random Fields}
 \newacronym{LSTM}{LSTM}{Long Short-Term Memory}
 \newacronym{MCHOG}{MCHOG}{Motion Contour image based Histogram Of Gradients}
 \newacronym{MLP}{MLP}{Multilayer Perceptron}
 \newacronym{MLE}{MLE}{Maximum Likelihood Estimation}
 \newacronym{PF}{PF}{Particle Filter}
 \newacronym{PHTM}{PHTM}{Probabilistic Hierarchical Trajectory Matching}
  \newacronym{ResNet}{ResNet}{Residual Networks}
 \newacronym{RGB}{RGB}{Red Green Blue}
 \newacronym{ROC}{ROC}{Receiver Operating Characteristic}
 \newacronym{RMSE}{RMSE}{Root Mean Squared Error}
 \newacronym{RNN}{RNN}{Recurrent Neural Network}
  \newacronym{SDD}{SDD}{Stanford Drone Dataset}
 \newacronym{SVM}{SVM}{Support Vector Machine}
 \newacronym{SLAM}{SLAM}{Simultaneous Localization and Mapping}
 \newacronym{SLDS}{SLDS}{Switching Linear Dynamical System} 
 \newacronym{TPR}{TPR}{True Positive Rate}
 \newacronym{V2I}{V2I}{Vehicle to Infrastructure}
 \newacronym{V2V}{V2V}{Vehicle to Vehicle}
 \newacronym{V2X}{V2X}{Communication between vehicles and anything}
 \newacronym{VR}{VR}{Virtual Reality}
 \newacronym{VAE}{VAE}{Variational Auto Encoders}
 \newacronym{VRU}{VRU}{Vulnerable Road User}
 \newacronym{UAV}{UAV}{Unmanned Aerial Vehicle}
 \newacronym{WHO}{WHO}{World Health Organization}

\title{\LARGE \bf
Scene Compliant Trajectory Forecast with Agent-Centric Spatio-Temporal Grids
}

\author{Daniela Ridel$^{1*}$, Nachiket Deo$^{2}$, Denis Wolf$^{1}$, and Mohan Trivedi$^{2}$% <-this % stops a space
\thanks{*This work was done when Daniela Ridel was a Fulbright Scholar at the Laboratory for Intelligent and Safe Automobiles}% <-this % stops a space
\thanks{$^{1}$Mobile Robotics Lab, University of Sao Paulo, Sao Carlos, SP 13566, Brazil
        {\tt\small danielaridel@usp.br denis@icmc.usp.br}}%
\thanks{$^{2}$Laboratory for Intelligent and Safe Automobiles, University of California, San Diego, CA 92092, USA
        {\tt\small ndeo@ucsd.edu trivedi@ucsd.edu}}%
}

\begin{document}

\maketitle
\thispagestyle{empty}
\pagestyle{empty}

%%%%%%%%%%%%%%%%%%%%%%%%%%%%%%%%%%%%%%%%%%%%%%%%%%%%%%%%%%%%%%%%%%%%%%%%%%%%%%%%
\begin{abstract}
Forecasting long-term human motion is a challenging task due to the non-linearity, multi-modality and inherent uncertainty in future trajectories. The underlying scene and past motion of agents can provide useful cues to predict their future motion. However, the heterogeneity of the two inputs poses a challenge for learning a joint representation of the scene and past trajectories. To address this challenge, we propose a model based on grid representations to forecast agent trajectories. We represent the past trajectories of agents using binary 2-D grids, and the underlying scene as a RGB \gls{BEV} image, with an agent-centric frame of reference. We encode the scene and past trajectories using convolutional layers and generate trajectory forecasts using a \gls{ConvLSTM} decoder. Results on the publicly available \gls{SDD} show that our model outperforms prior approaches and outputs realistic future trajectories that comply with scene structure and past motion.

\end{abstract}

%%%%%%%%%%%%%%%%%%%%%%%%%%%%%%%%%%%%%%%%%%%%%%%%%%%%%%%%%%%%%%%%%%%%%%%%%%%%%%%%

\section{Introduction}
The study of human motion has been broadly explored by several applications, as character animation, surveillance systems, traffic analysis, and autonomous driving. Humans' trajectories can be predicted to avoid collisions, to detect suspicious behavior, or even to monitor crowd flow. As part inherent of humans' motion they are constantly adapting their paths regarding goals they want to reach, obstacles they want to avoid, and rules they are obligated to obey.

When humans navigate in urban spaces, they might be walking, cycling, skating, or driving. These are just a few examples of types of transportation commonly used by humans. 

The type of transportation used by a person characterizes his/hers pattern of motion.  Therefore the person's trajectory is very correlated to scene, i.e. pedestrians usually walk on sidewalk, whereas drivers are expected to follow lanes. This suggests that scene semantic information is an important cue when dealing with different patterns of human motion.

Another meaningful information are the person's past positions, as they can help to understand the direction he/she is moving towards. A person's past trajectory can also restricts the space of probable future positions, as generally a person does not return to a preceding position. A high probable path for one person may have low probability to another just based on the direction both of them are walking towards.

Whether most of the current trajectory forecasting approaches extract features from a trajectory vector and concatenate with features extracted from the whole image to forecast trajectories, we establish an agent-centered spatio-time correlation between scene and past trajectory. Such spatio-time correlation is done by the usage of grids that represent both past trajectory and scene environment.  
The scene information here in the shape of a \gls{BEV} image can be obtained through an \gls{UAV}, smart city infrastructure (e\,.g. camera in a traffic light or at the top of a building), or even by the projection of a camera mounted on an \gls{AGV}.

We propose a method that is able to predict multi-modal diverse paths that have high correspondence between predicted trajectories and scene. 

The predicted trajectories avoid obstacles in most of the sequences (92\%), and they are also in compliance with path preferences from different agents, e\,.g. pedestrians prefer crosswalks whether cyclists prefer streets. We have outperformed state-of-the-art methods in the \gls{SDD} dataset, using a smaller number of predicted trajectories ($K=5$), achieving an ADE of 14.92 and a FDE of 27.97. Evaluating multi-modal trajectories is still an open problem as most of the current approaches are able to evaluate recall but fail into evaluating precision. Therefore we also propose a new measure to evaluate the percentage of all predicted positions (from all the predicted trajectories) that lie in paths, obstacles, and terrain.

\section{Related Studies}

A large body of literature has addressed the problem of human motion forecasting. For a more extensive review, we refer the reader to \cite{Shirazi2015IV,Ohnbar2016ITSC,Ridel2018,Rudenko2019Arxiv}. Here, we review deep learning models for motion forecasting. In particular, we focus on how these models encode the past motion and static scene context, and how they address the multi-modality of future motion.

\textbf{Encoding past motion:}
The past motion of agents is the simplest cue for forecasting their future motion. Past motion is typically represented using sequences of location co-ordinates obtained via detection and tracking. A majority of approaches encode such sequences using Recurrent Neural Networks (RNNs) such as Long Short-Term Memory (LSTMs) networks or Gated Recurrent Units (GRUs)  \cite{Alahi2016CVPRSocialLSTM,Zyner2019ITS,Amirian2019Arxiv,Gupta2018CVPRSocialGAN,Hasan2018Arxiv,Sadeghian2019CVPR, deo2018convolutional}. Alternatively, some approaches use temporal convolutional networks for encoding sequences of past locations \cite{Lee2017CVPRDesire,nikhil2018convolutional}, allowing for faster run-times. In addition to location co-ordinates, some approaches also incorporate auxiliary information such as the head pose of pedestrians \cite{Hasan2018Arxiv, Ridel2019IV} while encoding past motion. Many approaches jointly model the past motion of multiple agents in the scene to capture interaction between agents \cite{Alahi2016CVPRSocialLSTM,Liang2019CVPR,Lee2017CVPRDesire,Sadeghian2019CVPR,Amirian2019Arxiv,deo2018convolutional}. This is typically done by pooling the RNN states of individual agents in a \textit{social tensor} \cite{Alahi2016CVPRSocialLSTM, Lee2017CVPRDesire, deo2018convolutional}, using graph neural networks \cite{vemula2018social} or by modeling pairwise distances between agents along with max pooling \cite{Gupta2018CVPRSocialGAN, Sadeghian2019CVPR, Amirian2019Arxiv}.      

\textbf{Encoding the static scene context:}
Locations of static scene elements such as roads, side-walks, crosswalks, and obstacles such as buildings and foliage constrain the motion of agents, making them a useful cue for motion forecasting. Most recent approaches use Convolutional Neural Networks (CNNs) to encode the static scene context, either by applying the CNNs to bird's eye view images \cite{Sadeghian2019CVPR,Sadeghian2018ECCVCARNET,Lee2017CVPRDesire}, high fidelity maps \cite{cui2018Arxiv,chou2019Arxiv}, or LiDAR point cloud statistics in the bird's eye view \cite{Zeng2019CVPR,rhinehart2018r2p2}.       
Prior approaches use very different models for encoding the past motion and static scene context. This heterogeneity of the two inputs poses a challenge to learning a joint representation over them. Thus, we propose a model based on grid representations. We represent the past trajectories of the agents using binary 2-D grids, and the underlying scene as a RGB Birds-Eye View (BEV) image, with an agent-centric frame of reference. We encode the scene and past trajectories using convolutional layers and generate trajectory forecasts using a Convolutional LSTM (ConvLSTM) decoder. Closest to our approach is the model proposed by Li \cite{Li2017ICM}, which uses a ConvLSTM encoder-decoder trained on a grid based representation of past motion. However, unlike our model, they do not encode the static scene.

\textbf{Models incorporating multi-modality:}
An inherent difficulty in motion forecasting is its multi-modal nature. There are multiple plausible future trajectories at any given instant due to latent goals of agents and multiple paths to each goal. Regression based approaches for motion forecasting tend to average these modes, often leading to implausible forecasts. Prior works have addressed this challenge by learning one-to-many mappings. This is most commonly done by sampling generative models such as Generative Adversarial Networks (GANs) \cite{Gupta2018CVPRSocialGAN, Sadeghian2019CVPR,Amirian2019Arxiv}, Variational Autoencoders (VAEs) \cite{Lee2017CVPRDesire} and invertible models \cite{rhinehart2018r2p2}. Some approaches sample a stochastic policy obtained using imitation learning or inverse reinforcement learning \cite{Li2019CVPR, Deo2019ICRA}.  Other approaches learn mixture models \cite{cui2018Arxiv, Zyner2019ITS, deo2018convolutional, Deo2018IV}.

In this work, we output a fixed number of output trajectories, and use the 'best of k' prediction loss to train the model similar to \cite{Gupta2018CVPRSocialGAN, cui2018Arxiv}.

\section{Problem Formulation and Notation}
\textbf{Problem Formulation:}
Our model uses past trajectories of agents and scene to predict future positions.

A trajectory is defined as a sequence of $x,y$ positions in respect to time. $\tau_{h}$ represents an agent past trajectory until upon time $t$. At inference time, $t$ represents the last observation of an agent's position.

\begin{equation}
\tau_{h} = \left [ \textbf{x}^{t-t_{h}},\cdots,\textbf{x}^{t-1},\textbf{x}^t \right ]
\end{equation}

$\tau_{f}$ represents an agent future trajectory ranging from time $t+1$ to $t+ t_{f}$.
\begin{equation}
\tau_{f}  = \left [ \textbf{x}^{t+1},\textbf{x}^{t+2},\cdots, \textbf{x}^{t+t_{f}} \right ]
\end{equation}

where $\textbf{x} = \left [ x, y\right ]$, and $t_{h}$ and $t_{f}$ are, respectively, the number of past and future positions used as time window.
Given an agent past trajectory $\tau_h$ and scene information, we want to predict $\tau_{f}$. We denote $\tau_{f}$ the \gls{GT} trajectory, $\hat{\tau}_{f}$ a predicted trajectory, and $\hat{\tau}_f^k$ with $k \in \left \{  1, 2, ..., K\right \}$ each one of the $K$ predicted trajectories. %$\hat{\tau_{f}}$

We transform each trajectory $\tau_h$ to grid representation.%, enabling a spatial correlation between trajectory and scene.

\textbf{Trajectory Representation with Grids:}
For each trajectory $\tau_{h}$ we generate a  $N \times N$ Boolean grid with $t_{h}$ number of channels. Each grid channel is populated according to the trajectory $x$ and $y$ positions at each time step. That means each grid is zeroed and has an one on the corresponding $(x,y)$ time step position.

\textbf{Scene Representation with Grids:}
Scene is represented by a $N \times N$ grid with three channels. Each grid position stores the RGB pixel values of a \gls{BEV} map of the environment. Scene and trajectory are represented in the same spatial manner.

\textbf{Frame of Reference:} The frame of reference is centered at each agent being predicted at time $t$. For each trajectory we consider the $x$ and $y$ positions of the agent at time $t$ as $(0,0)$ adapting the past and future positions according to this reference. The grid representations are therefore centered at the agent position at time $t$.

\begin{figure*}
	\centering
	\includegraphics[width=1.0\textwidth]{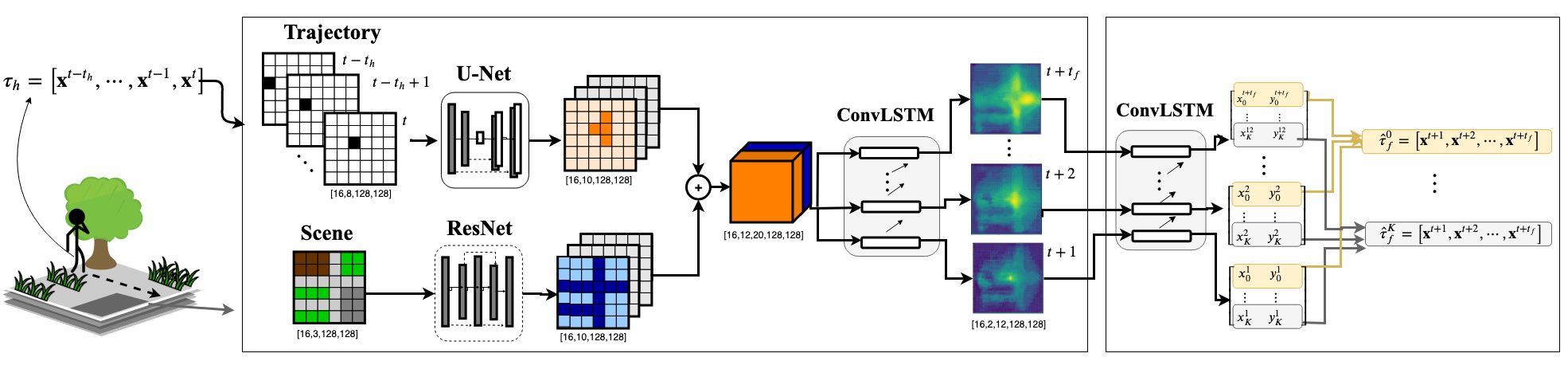}
	\caption[B]
	{\textbf{Proposed model for scene compliant trajectory forecasting with spatial grids}.  An U-Net with skip connections processes the trajectory grid and a ResNet processes the scene grid. The concatenation of the outputs from both U-Net and ResNet are used as input to the \gls{ConvLSTM} model that outputs the probability grids. The sampling module uses the generated probability grids to create $K$ possible trajectories $\hat{\tau}_f$.}
	
	\label{fig:diagram}
\end{figure*}

\section{Proposed Model}
Our network comprises two modules, Fig. \ref{fig:diagram}. The first one (probability grid generation) takes as input trajectory and scene and generates the probability grids. These grids store the probability of the agent being in each cell at a determined time-step. The second module (sampling) samples $K$ trajectories ($\hat{\tau}_f$) from the probability grids.

\subsection{Probability Grid Generation}
\subsubsection{Scene}
\gls{ResNet} \cite{he2016CVPRResnet} were used to process the scene. Such networks preserve specific features while also reasoning about global features of the scene. \gls{ResNet} is also useful to train because it can transform the loss search space into a smother function \cite{Li2018Nips}.

We pre-trained the ResNet using satellites images from the \gls{ISPRS} \cite{rottensteiner2012isprs} Potsdam dataset, to solve the semantic segmentation problem using cross entropy loss. The \gls{ISPRS} provides semantic labels for 6 classes (impervious surfaces, buildings, low vegetation, tree, car, and clutter/background) in satellite's images. Such dataset does not have a specific label for sidewalks, as we consider that this is an important information for our path prediction problem we hand labeled some images from the training set of \gls{SDD} \cite{Robicquet2016ECCV} dataset and further trained the model to semantic segment such images.

\subsubsection{Trajectory}
 Both trajectory and scene are represented in grids. For each past trajectory $x,y$ position we generate a zeroed Boolean grid, setting the respective grid $x,y$ position to one. We need $t_h$ grids to represent the pedestrian  history of positions, whether for scene we only need three grids to represent the RGB values of the scene \gls{BEV} image.

The past trajectory grid is processed by an U-Net \cite{Ronneberger2015LNCSUnet} with skip connections. The choice of such architecture was made because as we are forecasting slow and fast moving agents in the same network we had to make sure all grid positions would be convoluted to encode the trajectory. As stated in prior work, for trajectory prediction, the most recent positions of an agent have more influence in his/her future positions than older positions. Such past trajectory information is commonly useful to learn the orientation as in most cases the agents do not return to positions they have already been to, except the cases where they are stopped. 
 
\subsubsection{ConvLSTM}

The prediction of long-term future trajectories tend to be more challenging and strongly relate to scene. Such future trajectories can be more robust and comprise curves to avoid obstacles. Because of this extra challenge we use \gls{ConvLSTM}s instead of the simple convolution networks, such architecture can learn more robust trajectories by deciding which features it should use from the prior \gls{LSTM} cell and the current input, learning what it can forget or remember in order to reason about the future. 
We use weighted cross entropy loss to train the grid generation (first module in Fig. \ref{fig:diagram}).
Our model outputs $t_{f}$ grids with probable agent future positions.

\begin{figure*}
    \centering
    \subfigure{\includegraphics[width=0.22\textwidth]{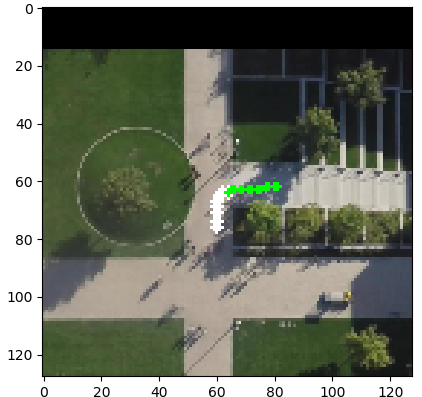}\includegraphics[width=0.65\textwidth]{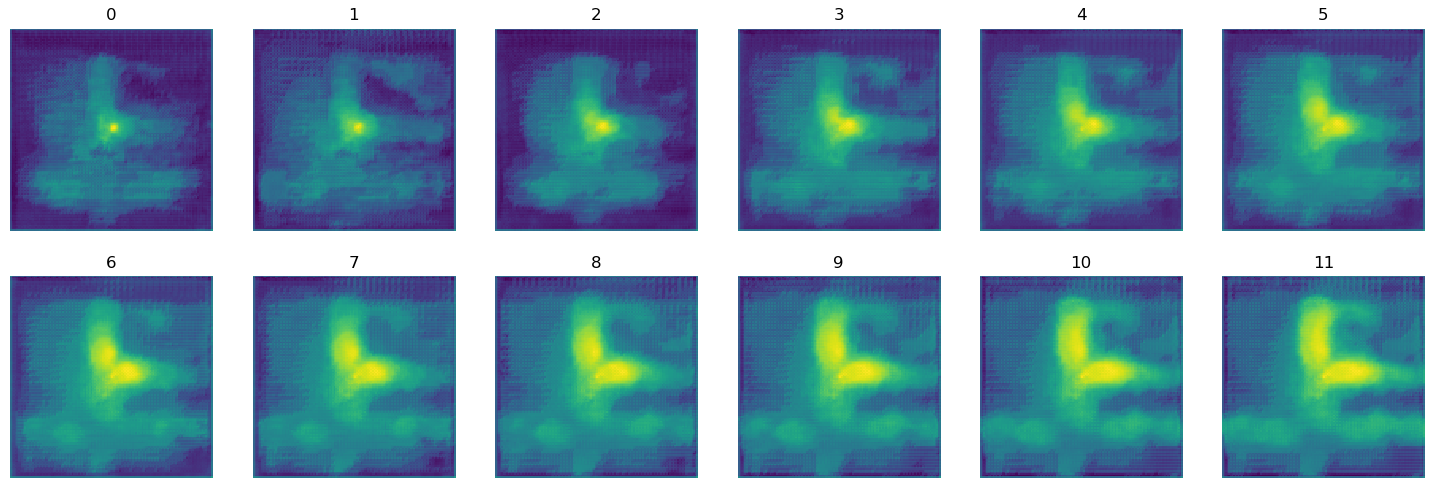}}%
    \caption{\textbf{Qualitative example of probability grids generated by the proposed method}. The BEV image on the left contains a scenario where an agent's past motion is represented in white, and the ground truth future motion is represented in green. The set of images on the right contains the $t_f=12$ grids generated by the proposed approach according to the respective BEV image on the left. Each grid corresponds to one predicted time-step. Each grid's cell store the probability of the agent occupy that cell at that time-step. Closer to yellow higher the probability. \textbf{Left image:} The past trajectory (white) ends in an intersection where the agent can choose between walking towards the top or to turn to the right (according to the viewer perspective). The GT trajectory (green) shows that in that specific case the agent decided to turn right. \textbf{Right image:} In the grids 0 and 1 the most probable cells are closer to the center of the grid. from grid 2 to grid 11 the generated grids are clearly assigning higher probabilities to the cells that correspond to both top and right paths. By reasoning about the past trajectory the model is able to distinguish different future trajectories in the same given scene.}
    \label{fig:grids}
\end{figure*}
 
\subsection{Sampling}

In order to compare our method with prior approaches we have to extract diverse and cohesive trajectories from the probability grids computed by the first module. The sampling step, second module in Fig. \ref{fig:diagram}, receives as input the grid maps and outputs $K$ predicted trajectories,  $\hat{\tau}_f^k$ with $k \in \left \{  1, 2, ..., K\right \}$, using \gls{ADE} to compute the loss. Figure  \ref{fig:grids} displays an example of $t_f = 12$ grids generated by our first module, where from grid 1 to grid 12 the probabilities shift from the center of the image towards the most probable paths/cells.

Variety loss have been used in prior work \cite{Gupta2018CVPRSocialGAN,Lee2017CVPRDesire,Sadeghian2019CVPR} with $K$ value, usually, being 5, 10, or 20. Such value means the number of outputs the model is trained to generate. All generated trajectories are then compared with the \gls{GT} trajectory and only the best loss is backward during training.

\subsection{Implementation Details}
All implementations were made using Pytorch. We applied a random rotation to all grids during training and we random shuffled the batches at every epoch. We used Adam optimizer \cite{Kingma2014Adam} with a scheduler to reduce the learning rate on plateau in case the loss did not improve during the last 4 epochs.

\subsubsection{Grid generation}
 We used a grid size of $N = 128$ because it allows us to fit most of the trajectories after down sampling them by a factor of ten, $t_h=8$ (3.2 secs), and $t_f=12$ (4.8 secs) as previously used in related work \cite{Sadeghian2019CVPR}. The U-Net and ResNet architectures were adapted from \cite{Isola2017CVPR}\footnote{\footnotesize{\url{https://github.com/phillipi/pix2pix}}}. We removed the \textit{tanh} function in the last layer. We used seven blocks of U-Net with skip connections and ResNet with nine blocks. All convolutions in both architectures follow a Convolution-BatchNorm-ReLU or a Convolution-BatchNorm-Dropout-ReLU sequence, and use $4 \times 4$ spatial filters with stride $2$. The convolutions down-sample and up-sample by a factor of two. The weights were initialized from a Gaussian distribution with mean zero and standard deviation 0.02. All ReLUs used in down sample are LeakyReLUs with slope = 0.2. Dropout rate was 0.5.
The \gls{ConvLSTM}\footnote{\footnotesize{\url{https://github.com/ndrplz/ConvLSTM_pytorch}}} architecture  have one layer with input dimension of $20$, hidden state with 16 channels, and kernel size of $(11,11)$. The last layer is a 2D convolution with input 16 and output 2, i\,.e. for each cell we gather the chance of it being part of the agent's trajectory or not.
\subsubsection{Sampling}
The ConvLSTM used in the sampling step has the same hyper parameters of the ConvLSTM used for grid generation, however the last layer was replaced for a fully connected layer. The network was trained using mADE loss.

\section{Experimental Analysis}
\label{sec4_experiments}

We conduct experiments on \gls{SDD} \cite{Robicquet2016ECCV} dataset to quantitatively and qualitatively evaluate  our approach. We use the standard train, validation and test split available in TrajNet \footnote{\url{http://trajnet.stanford.edu/}}. The \gls{SDD} dataset comprises different scenarios captured by a drone's camera. For each scene several trajectories are pixel-wise labeled. Those trajectories comprise diverse agents (pedestrians, cyclists, skaters, cars, buses, and carts). 
 
We excluded lost positions from the sets of trajectories. A new ID was created when the agent re-appeared in the scene. 

\subsection{Baselines and Metrics}
Efficiently evaluating multi-modal trajectories is still a open problem. 
Most of current approaches using best-of-$K$ (or variety loss), compare each one of the $K$ predicted trajectories with the \gls{GT} trajectory, and consider as the result the predicted path that achieved closest distance to the \gls{GT}. 
A current problem of such approach is that it is able to measure recall but it fails into capturing the precision. To evaluate the precision of our $K$ predicted trajectories, we also use a Correspondence to Scene metric that gives us the percentage of predicted trajectory points that lies on paths, terrain, and obstacles. As such trajectories should not pass through obstacles, such measurement access the precision of the proposed approach.

Given the \gls{GT} trajectory $\tau_f$ and the K predicted trajectories $\hat{\tau}_f^k$ with $k \in \left \{  1, 2, ..., K\right \}$, we compute three metrics to evaluate the proposed method.

\textbf{Minimum Average Displacement Error (mADE):} Minimum value among the average distance between each predicted trajectory and GT.

\begin{equation}
   mADE = \min_{k \in \{1,2, \dots, K\}}\frac{1}{T}\sum_{t=1}^{T}\left \| \tau_t - \hat{\tau}^{(k)}_{t}\right \|_{2},
\end{equation}
where $\tau_t$ is the GT trajectory position at time $t$ and $\hat{\tau}_t^{(k)}$ is the position of predicted trajectory $k$ at time $t$.

\begin{figure}
    \centering
    \subfigure[]{\includegraphics[width=.45\columnwidth]{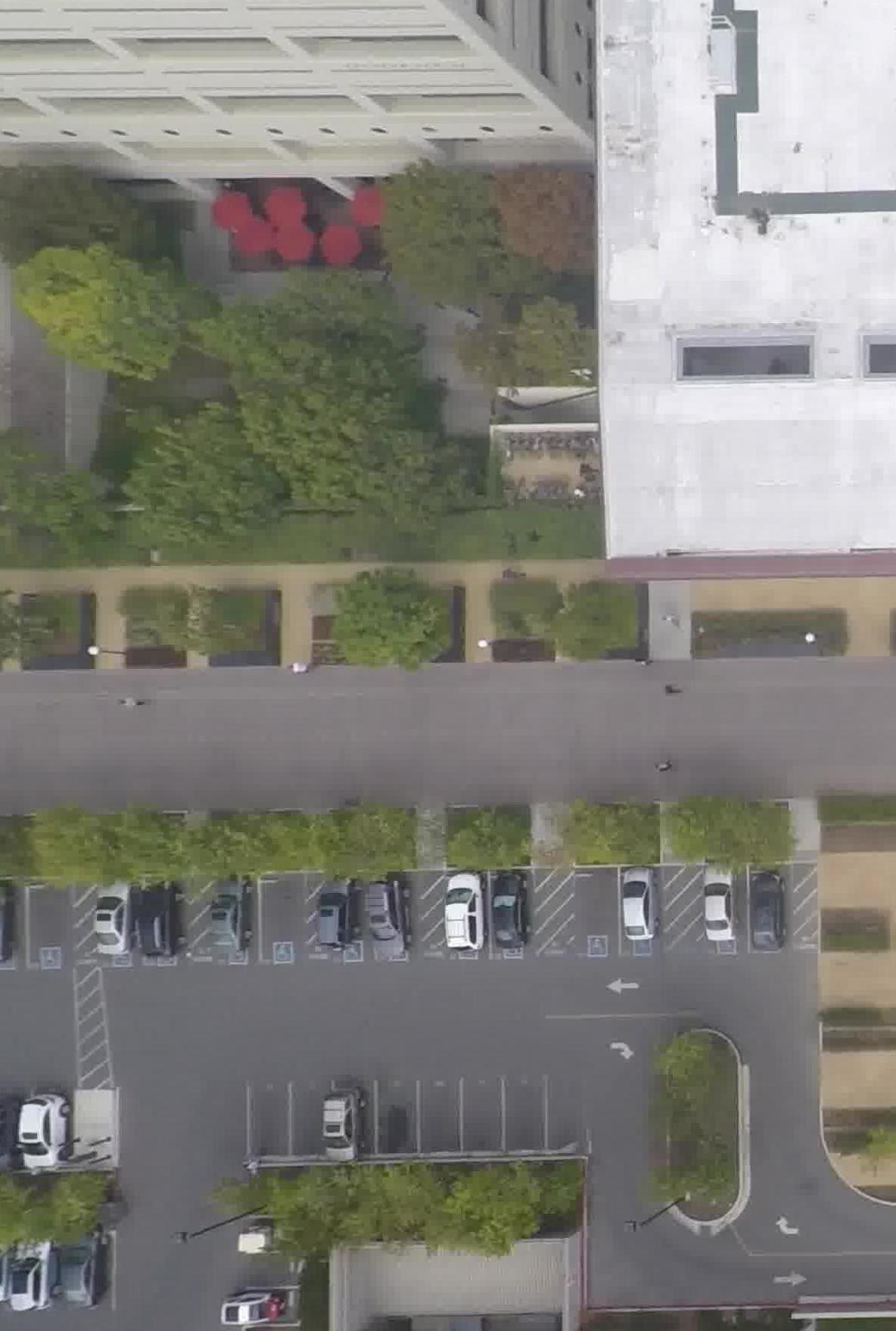}}
    \quad
    \subfigure[]{      \includegraphics[width=.45\columnwidth]{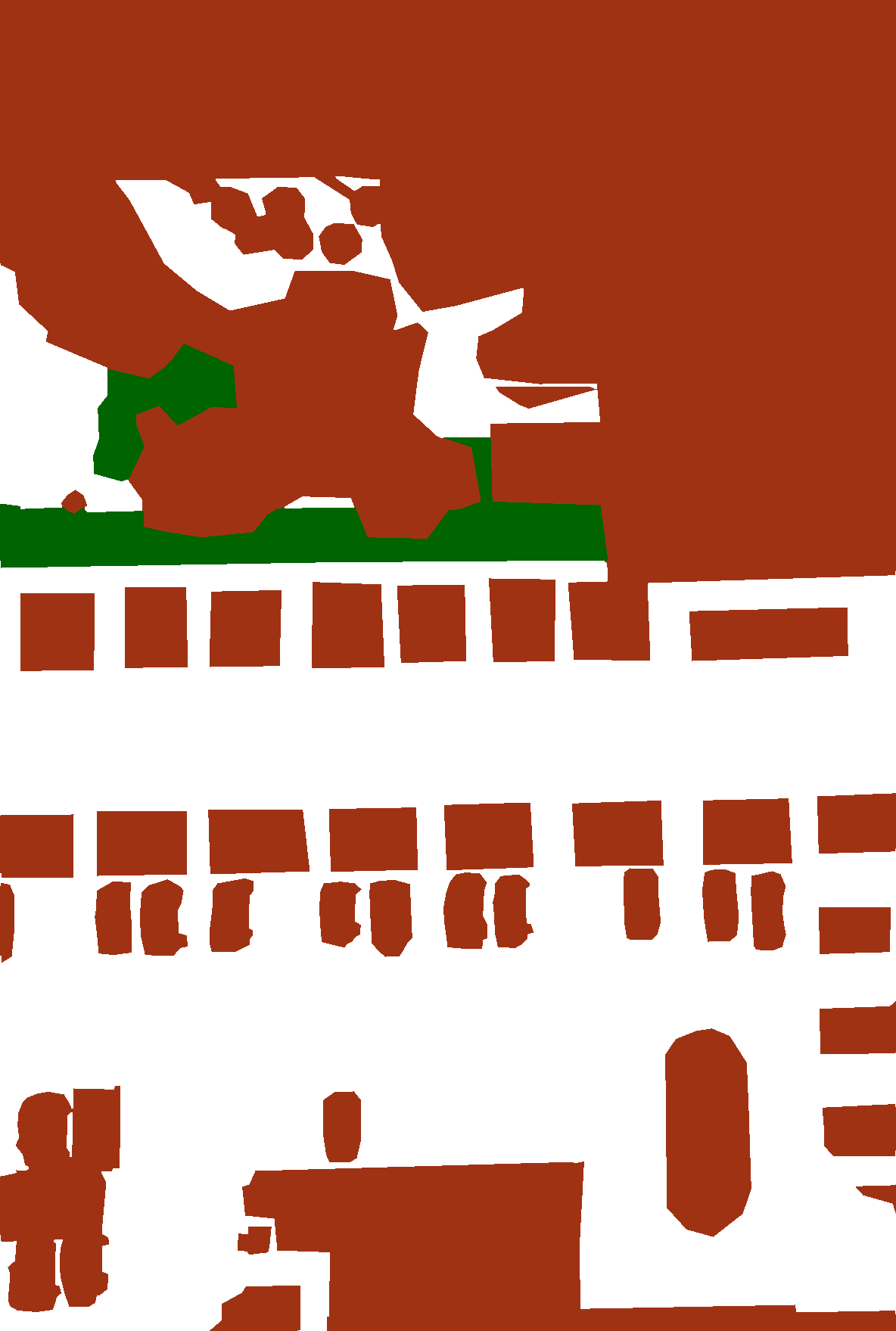}}
    \caption{\textbf{Illustration of manually accomplished semantic labeling.} (a) BEV image (Nexus 5) from SDD dataset, (b) Semantic labeled image,  obstacle (red), terrain (green), and path (white).}
    \label{fig:labelCS}
\end{figure}

\textbf{Minimum Final Displacement Error (mFDE):} Minimum final displacement error between each predicted trajectory final point and the GT final point.

\begin{equation}
   mFDE = \min_{k \in \{1,2, \dots, K\}}\left \| \tau_T - \hat{\tau}^{(k)}_{T}\right \|_{2},
\end{equation}

and \textbf{\gls{CS}:} For each image in the testing set we hand labeled the pixels into path (sidewalk, street), terrain (grass, ground), or obstacle (trees, cars, buildings), e\,.g. Fig.\ref{fig:labelCS}. For each one of the $K$ predicted trajectories, we match in the labeled image if each point in the predicted trajectory lies in a pixel that corresponds to a path, terrain or obstacle. We sum up all the points that lies in each category and then we divided by the total number of points, The result is the percentage of points that lies in path, terrain and obstacle.

\subsection{Results}
\begin{figure*}
    \centering
    \subfigure[]{      \includegraphics[width=0.22\textwidth]{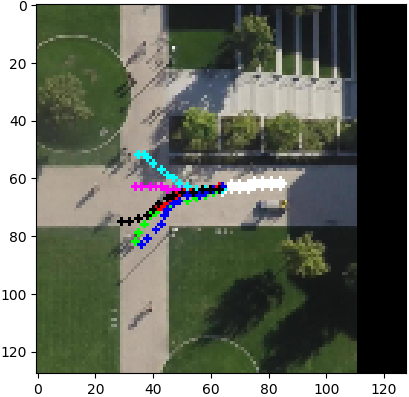}}
    \quad
    \subfigure[]{      \includegraphics[width=0.22\textwidth]{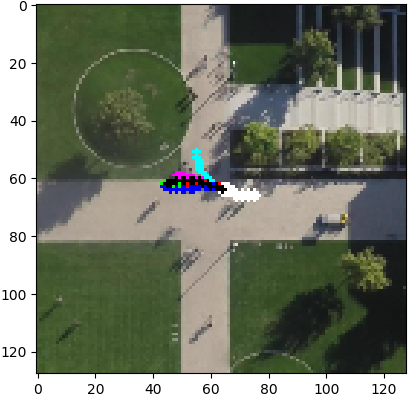}}
    \quad
    \subfigure[]{\includegraphics[width=0.22\textwidth]{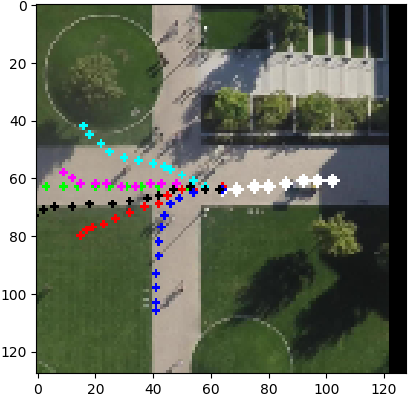}}
    \quad
    \subfigure[]{\includegraphics[width=0.22\textwidth]{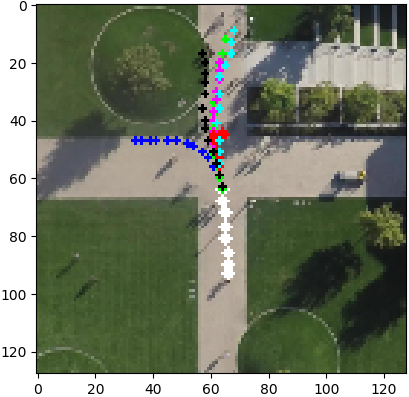}}
    \quad
    \subfigure[]{      \includegraphics[width=0.22\textwidth]{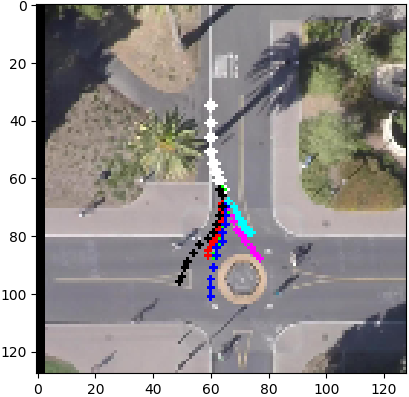}}
    \quad
    \subfigure[]{      \includegraphics[width=0.22\textwidth]{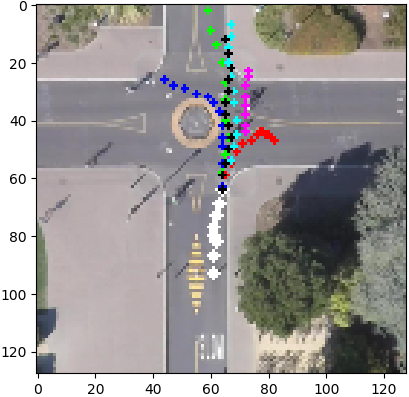}}
    \quad
    \subfigure[]{      \includegraphics[width=0.22\textwidth]{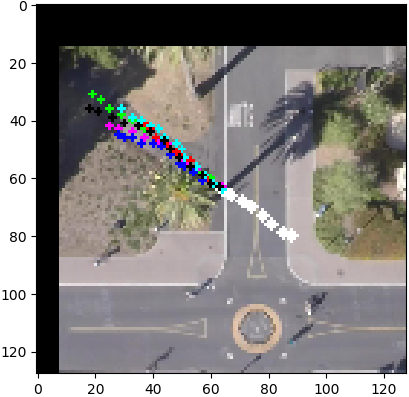}}
    \quad
    \subfigure[]{\includegraphics[width=0.22\textwidth]{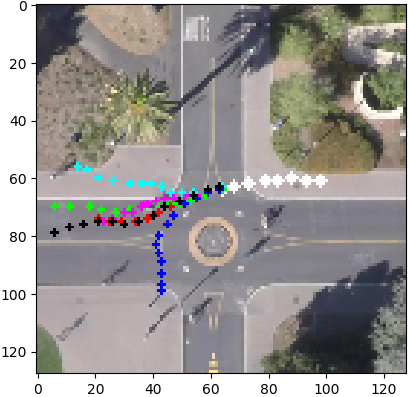}}
    \quad
    \subfigure[]{      \includegraphics[width=0.22\textwidth]{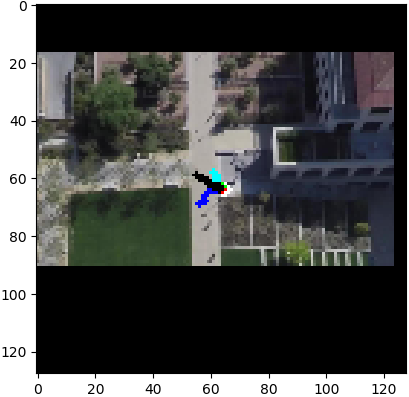}}
    \quad
    \subfigure[]{\includegraphics[width=0.22\textwidth]{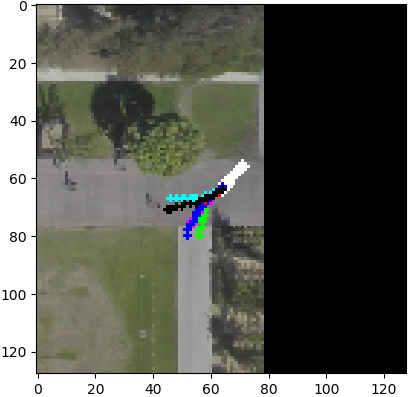}}
    \quad
    \subfigure[]{      \includegraphics[width=0.22\textwidth]{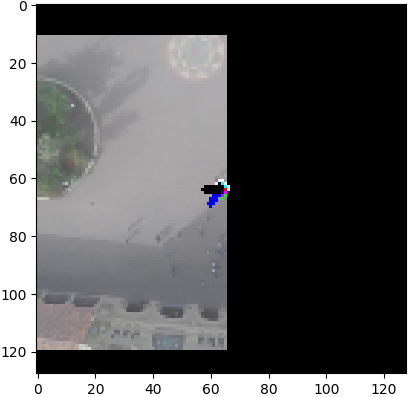}}
    \quad
    \subfigure[]{\includegraphics[width=0.22\textwidth]{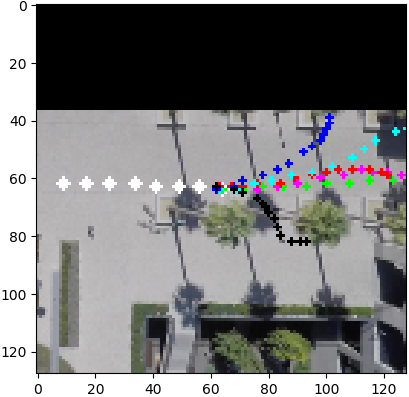}}

    \caption{\textbf{Qualitative results of the proposed method}. White represents past trajectory, green represents future trajectory (\gls{GT}) and other colors (light blue, dark blue, black, red, and magenta) represent the five predicted trajectories. The scene information tends to bound the possible paths and the past trajectory define the most probable trajectories based in the paths delimited by the scene encoding. In (a) according to the past trajectory direction there are three visually well delimited paths: left, top, and down (according to the viewer perspective). The predicted trajectories are in compliance with the three possibilities, in (b) despite the possible paths are the same as Fig. (a), the past trajectory is moving towards the top. All the predicted trajectories go towards the left and top path, and the down path is disregarded, in (c) the predicted trajectories are more spread trying to cover a more broad number of possibilities, in (d) there is one predicted trajectory turning towards the left path, all other predicted trajectories are going towards the top, following the GT trajectory, in (e) all predicted trajectories avoid the roundabout even with the last point on the past trajectory going towards it, in (f) the predicted trajectories belong to very feasible paths (top, left, and right), in (g) there is a very narrow paved path in the middle of the top-left terrain. The path can be seen in (Fig. e and h). All predicted trajectories stick together through the narrow path, in (h) the predicted trajectory in light blue continues the past trajectory. The dark blue trajectory assume the possibility of the agent crossing the street, and the other predicted trajectories go in GT direction, in (i) the agent is stopped. Three trajectories try to guess possible trajectories in case the agent decide to move, in (j) there is two possible visually well delimited paths, left and down. All predicted trajectories are situated in both paths, in (k) the agent is stopped and the predicted trajectories are only towards the portion left of the grid, that has semantic information, avoiding the dark side, and in (l) There is not a well delimited path. All predicted paths go towards the top, bottom and right. }
    \label{fig:results_predicted_trajs}
\end{figure*}

We directly report the results from \cite{Sadeghian2019CVPR} in Tab. \ref{tab:resultssophie}. Our method with $K = 5$ outperformed prior state-of-the-art (using $K=20$), in both ADE ($14.92$) and FDE ($27.97$) metrics. 
In Fig. \ref{fig:results_predicted_trajs} we present some qualitative results. The points in white are the agent's past positions; the green points are the \gls{GT} future points; the points in light blue, dark blue, black, red, and magenta are the predicted trajectories. In several scenes where the agent is walking towards an intersection Fig. \ref{fig:results_predicted_trajs}a, b, c, d, e, f, h, j; a roundabout Fig. \ref{fig:results_predicted_trajs}e, f, h; a big free area Fig. \ref{fig:results_predicted_trajs}l; a straight path Fig. \ref{fig:results_predicted_trajs}g; or he/she is stopped Fig. \ref{fig:results_predicted_trajs}k, i; the predicted trajectories are highly correlated to scene. Nearly all predicted trajectories were contouring the roundabout and were plausible extensions of the past trajectory. Usually when the agent is in the middle of a big free area there is a pattern of having at least tree clear predicted paths, e\,.g. Fig. \ref{fig:results_predicted_trajs}l. In Fig. \ref{fig:results_predicted_trajs}g, the path is very narrow and all predicted trajectories are very close to each other. In many figures it is possible to see the compliance with scene. In Fig. \ref{fig:results_predicted_trajs}j, given the past trajectory there are two possible paths left or down (according to the viewer perspective) and all the predicted trajectories are lying in those two possibilities. In Fig. \ref{fig:results_predicted_trajs}b, despite there are three paths (top, left, and down), the past trajectory is inclined towards the top path, so the network focused the predicted trajectories in the top and the left path.

\begin{figure*}
    \centering
    \subfigure[]{\includegraphics[width=0.31\textwidth]{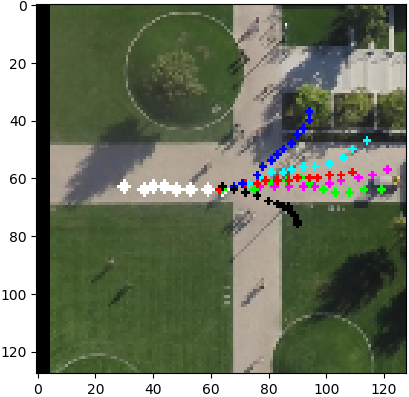}}
    \quad
    \subfigure[]{      \includegraphics[width=0.31\textwidth]{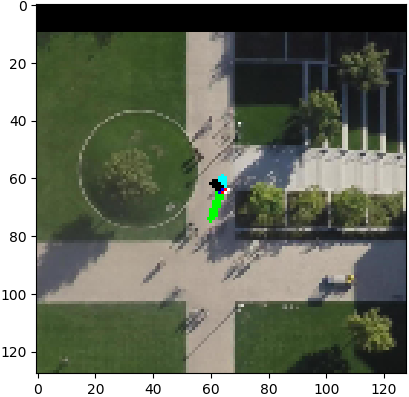}}
    \quad
    \subfigure[]{      \includegraphics[width=0.31\textwidth]{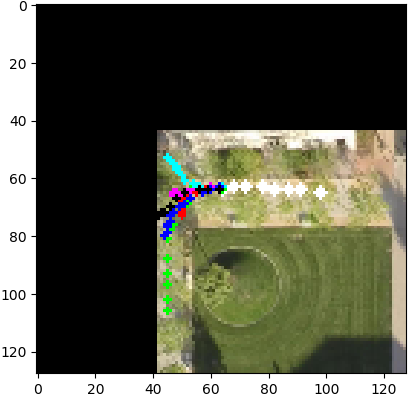}}
    \caption{\textbf{Illustrative cases where network performance can be improved.} In case (a) the blue trajectory terminates on an obstacle (tree), in (b) when an agent starts moving suddenly, orientation of the future trajectory may be inaccurate, and in (c) length of the future trajectory may be inaccurate possibly due to trees covering the path. White represents past trajectory, green future trajectory (\gls{GT}) and other colors (light blue, dark blue, black, red, and magenta) represent the $K=5$ predicted trajectories.}
    \label{fig:errors}
\end{figure*}

In Fig. \ref{fig:errors} we point some of the cases where our model fails. In Fig. \ref{fig:errors}a the last position of the dark blue trajectory lies in the top of a tree, in that case the network could not distinguish the tree from the grass.
Scenarios where the agent is stopped in the $\tau_h$ and start to walk in the $\tau_f$ are still a challenge. Such challenge arises due to the lack of information regarding orientation in $\tau_h$, as shown in Fig. \ref{fig:errors}b. Despite the predicted trajectories in Fig. \ref{fig:errors}b are feasible, none of them match the GT trajectory (green). In Fig. \ref{fig:errors}c the network predicted a shorter trajectory than the \gls{GT}, this probably happened because there are some trees' tops projected in the path due to the camera view perspective.

We display the results for the CS measure in Table \ref{tab:resultscs}. One observation from such table is that there are some points in the GT that lies in obstacles. In the SDD data there are scenarios where pedestrians are partially walking inside buildings, and as we hand labeled buildings as obstacles the trajectories' points will be computed as obstacles even if an indoor path existed. In order to fully understand the $\%$ on obstacles we have to look to both GT and ours results.
\begin{table*}
\normalsize
\caption{Quantitative comparative performance analysis of the proposed approach. Note, results indicate significant improvement over selected state-of-the-art approaches as measured by two commonly used metrics (ADE and FDE).}
\begin{center}
\begin{tabular}{|c|p{1.5cm}|p{1.1cm}|p{1.1cm}|p{1.1cm}|p{1.5cm}|p{1.7cm}|p{1.7cm}|p{1.7cm}|}\hline
\textbf{Method}&Social GAN \cite{Gupta2018CVPRSocialGAN}&Sophie \cite{Sadeghian2019CVPR}&Desire \cite{Lee2017CVPRDesire}&\textbf{Ours}&Linear Regressor&Social Forces \cite{Yamaguchi2011CVPRSocialForces}&Social LSTM \cite{Alahi2016CVPRSocialLSTM}&CAR-NET \cite{Sadeghian2018ECCVCARNET}\\\hline%\midrule
\textbf{\emph{K}} (train and test)&20&20&5&5&1&1&1&1 \\\hline%\midrule
\textbf{\emph{ADE}} \small{(pixels)}&27.24&16.27&19.25& \textbf{14.92}&37.11&36.48&31.19&25.72 \\\hline%\midrule
\textbf{\emph{FDE}}\small{(pixels)}&41.44&29.38&34.05&\textbf{27.97}&63.51&58.14&56.97& 51.8 \\\hline%\midrule

 \hline
\end{tabular}
\end{center}
\label{tab:resultssophie}
\end{table*}

\begin{table*}
\normalsize
\caption{Quantitative performance results of forecasted trajectories as compared against the ground truth.}
\begin{center}
\begin{tabular}{|c|p{0.5cm}|p{1.7cm}|p{2cm}|p{2.5cm}|p{3.5cm}|}\hline
\textbf{Method} & \textbf{K} & \textbf{\% on path} & \textbf{\% on terrain} & \textbf{\% on obstacles} & \textbf{\% out of the image}\\\hline%\midrule 

Ours & 5 &  86.35\%& 5.74\% & 7.89\% & 0.01\%\\ 
\textbf{\gls{GT}} & 1 & 87.88\% & 4.95\% & 7.16\% & 0.0\%\\ \hline

\end{tabular}
\end{center}
\label{tab:resultscs}
\end{table*}

In general the proposed approach was able to generate diverse trajectories that comply with pedestrian past trajectory and scene. To deal with trajectories represented in image space is a non trivial task as the size of the grid directly implies on the maximum trajectory size that can be represented in such structure.

\section{Concluding Remarks}
In this work we have explored the prediction of multi-modal trajectories by using spatio-time compliant representations for both scene and trajectory with agent-centric grids. U-Net and ResNet were used to, respectively, encode trajectory and scene. \gls{ConvLSTM}s were used to generate probability grids and to sample trajectories. Our quantitative results on SDD dataset achieved state-of-the-art performance and qualitative results show that the predicted trajectories were in conformity with past trajectory, compliant to scene, and diverse. Future work can extend this model by exploring different sampling techniques, the usage of information regarding agent's interaction, and also the prediction of multiple agents' trajectories in parallel.

%%%%%%%%%%%%%%%%%%%%%%%%%%%%%%%%%%%%%%%%%%

\section*{ACKNOWLEDGMENT}
The authors are grateful for the generous and continuing support of the research by sponsors and team members of the UCSD LISA and CVRR labs. The authors also thanks Fulbright and CAPES for the financial support in this research. This study was financed in part by the Coordena\c{c}\~{a}o de Aperfei\c{c}oamento de Pessoal de N\'{i}vel Superior - Brasil (CAPES) - Finance Code 001.

%%%%%%%%%%%%%%%%%%%%%%%%%%%%%%%%%%%%%%%%%%%%%%%%%%%%%%%%%%%%%%%%%%%%%%%%%%%%%%%%

\bibliographystyle{IEEEtran}%IEEEtranS
\bibliography{egbib}

\end{document}